\documentclass[10pt,twocolumn,letterpaper]{article}
 
\usepackage{framed}
\usepackage{cvpr}
\usepackage{times}
\usepackage{epsfig}
\usepackage{graphicx}
\usepackage{amsmath}
\usepackage{amssymb}

\usepackage[pagebackref=true,breaklinks=true,letterpaper=true,colorlinks,bookmarks=false]{hyperref}

\cvprfinalcopy 

\def\cvprPaperID{1705} 

\ifcvprfinal\fi
\begin{document}

\title{Fisher Vectors Derived from Hybrid Gaussian-Laplacian\\ Mixture Models for Image Annotation}

\author{
Benjamin Klein, Guy Lev, Gil Sadeh,   and Lior Wolf\\
 The Blavatnik School of Computer Science\\
Tel Aviv University\\
{\tt\small beni.klein@gmail.com, levguy@gmail.com, gilsadeh@gmail.com, wolf@cs.tau.ac.il}
}
 	
\maketitle

\begin{abstract}
In the traditional object recognition pipeline, descriptors are densely sampled over an image, pooled into a high dimensional non-linear representation and then passed to a classifier. In recent years, Fisher Vectors have proven empirically to be the leading representation for a large variety of applications. The Fisher Vector is typically taken as the gradients of the log-likelihood of descriptors, with respect to the parameters of a Gaussian Mixture Model (GMM).
Motivated by the assumption that different distributions should be applied for different datasets, we present two other Mixture Models and derived their Expectation-Maximization and Fisher Vector expressions. The first  is a Laplacian Mixture Model (LMM), which is based on the Laplacian distribution. The second Mixture Model presented is a Hybrid Gaussian-Laplacian Mixture Model (HGLMM) which is based on a weighted geometric mean of the Gaussian and Laplacian distribution. An interesting property of the Expectation-Maximization algorithm for the latter is that in the maximization step, each dimension in each component is chosen to be either a Gaussian or a Laplacian.
Finally, by using the new Fisher Vectors derived from HGLMMs, we achieve state-of-the-art results for both the image annotation and the image search by a sentence tasks. The additional task of caption synthesis given a query image is then addressed by feeding the projected HGLMM Fisher Vectors to a Recurrent Neural Network.
\end{abstract}

\section{Introduction}

The standard pipeline of object recognition is usually comprised of three main steps: The first is extracting local descriptors. For example, one of the most widely used local descriptor for object recognition in images is the SIFT descriptor~\cite{SIFT}, which is often extracted densely over the image/video. In the second step, the local descriptors are pooled into a representation of a single vector by a nonlinear transformation. For example, the Bag of Words (BoW) representation, which draws inspiration from the text retrieval community~\cite{BOW} is one of the most basic and familiar pooling techniques. In the last step, the single vector representation is usually passed to a classifier that is trained by a suitable machine learning algorithm, e.g., SVM.

This paper focuses on the pooling step. There are many pooling techniques~\cite{BOW,VLAD}. However, the leading pooling technique in recent years is the Fisher Vector~\cite{perronnin2007fisher}, which provided state-of-the-art results on many different applications~\cite{simonyan2013fisher,peng2014action,chatfield2011devil,perronnin2010large}. In all of these contributions, the Fisher Vector of a set of local descriptors is obtained as a concatenation of gradients of the log-likelihood of the descriptors in the set with respect to the parameters of a Gaussian Mixture Model that was fitted on a training set in an unsupervised manner.

Many different improvements were suggested for the Fisher Vector~\cite{perronnin2010improving,sydorovdeep,simonyan2013deep} but all of them are in the context of the Gaussian Mixture Model. In~\cite{jia2011heavy}, Jia et al. showed empirically that the statistics of gradient based image descriptors, such as SIFT~\cite{SIFT}, often follow a heavy-tailed distribution which suggests that a Gaussian distribution does not capture well the descriptors' distribution, and that the Euclidean distance is not a suitable distance.  They advocate for the selection of a distance measure according to the appropriate probabilistic model that fits the distribution of the empirical data, and show that for the application of SIFT feature matching, a significant improvement is obtained by using the Laplacian distribution and the $L1$ distance instead of the Euclidean distance. Motivated by their findings, this paper presents and evaluates new variants of Fisher Vectors that are based on the Laplacian distribution.

By using the common assumption in the Fisher Vector that the covariance matrix is a diagonal one, we define the multivariate Laplacian distribution and the Laplacian Mixture Model (LMM). We explain how to fit a LMM by deriving the Expectation-Maximization (EM) equations and supply the Fisher Vector definition for this model. Similar to~\cite{perronnin2007fisher}, we approximate the diagonal of the Fisher Information Matrix in order to normalize the dynamic range of the different dimensions in the Fisher Vector variant presented.

In order to gain the benefits of the two distributions in a single model, we define a new distribution, the Hybrid Gaussian-Laplacian distribution, which can be seen as a weighted geometric mean of the Gaussian and Laplacian distributions. As before, we define the Hybrid Gaussian-Laplacian Mixture Model (HGLMM), derive the EM equations for fitting a HGLMM model, derive the Fisher Vector definition and approximate the diagonal of the Fisher Information Matrix. Although the distribution of each dimension in each component is a weighted geometric mean of a Gaussian and a Laplacian distribution, we show that in the EM algorithm, there is a sharp binary choice between the Gaussian and the Laplacian distributions that results directly from the Maximization-Step derivations.

We employ the new variants of the Fisher Vectors for tasks that match texts with images. In our experiments, the images are represented by either the overfeat~\cite{sermanet2013overfeat} or the VGG~\cite{oxford} Convolutional Neural Network as a single vector. The text is represented as a set of vectors obtained by the word2vec~\cite{mikolov2013distributed} method. This set is converted to a Fisher Vector based on one of the distributions: GMM, LMM, or HGLMM. Text to image matching is done using the Canonical Correlations Analysis algorithm~\cite{hotelling1936relations}. This combination of methods proves to be extremely potent. 

\section{Previous work}
\label{sec:prev}
\paragraph{Fisher Vectors} The Fisher Vector representation was first introduced in~\cite{perronnin2007fisher} and has been used successfully in various contexts. Since its introduction, many improvements have been suggested which have dramatically enhanced its performance. Some of the most widely used improvements were introduced by Perronnin et al.~\cite{perronnin2010large}. The first improvement is to apply an element-wise power normalization function, $f(z)=sign(z)|z|^{\alpha}$ where $0\leq\alpha\leq 1$ is a parameter of the normalization. The second improvement is to apply a L2 normalization on the Fisher Vector after applying the power normalization function. By applying these two operations~\cite{perronnin2010large} achieved state-of-the-art accuracy on CalTech 256 and showed superiority over the traditional Bag of Words (BoW) model. Sydorov et. al~\cite{sydorovdeep} introduced Deep Fisher Kernels which is a successful attempt of combining two methods with a significant impact on object recognition - Fisher Kernels and Deep Learning. In the traditional pipeline of object recognition, the first two steps (extracting local descriptors and pooling) are done in an unsupervised manner, independently of the task. Only in the last step, when learning the classifier is the nature of the task taken into account. One of the advantages of Deep Learning is that the entire pipeline is optimized for the task - including the layers that are in charge of extracting descriptors and pooling. Sydorov et. al formulated the traditional pipeline of using Fisher kernel with a SVM classifier as a single multi-layer feed forward network. Therefore, both the GMM parameters and the weight vector of the classifiers are tuned according to the nature of the specific task. Simonyan et al.~\cite{simonyan2013deep} were motivated by the recent success of Convolutional Neural Networks (CNN)~\cite{lecun1998gradient} and proposed a version of the state-of-the-art Fisher Vector image encoding that can be stacked in multiple layers. Their version obtained competitive results with CNNs on the ILSVRC-2010 dataset. Furthermore, they demonstrated that their Fisher Vector version and CNNs representations are complementary and by combining the two, they achieved a significantly improved accuracy. Our method differs from previous work  in  that previous work concentrated on Fisher Vectors derived from the Gaussian Mixtue Model. In our work, we are deriving the Fisher vector for other distributions and show that for some tasks, the Fisher Vector variants that are based on  LMM or HGLMM can surpass the performance achieved by the conventional Fisher Vector.

\paragraph{Image Annotation and Image Search} There has been a recent growing interest in methods that can bridge between the domains of vision and NLP. The works of~\cite{mitchell2012midge,kuznetsova2012collective,li2011composing} have focused on generating novel descriptive sentences for a query image. Kulkarni et al.~\cite{kulkarni2011baby} suggested a system that generates a descriptive text (not from a fixed set) for a given query image. Their pipeline contains the following steps; First, object detectors find candidate objects in an input image. The candidates objects are passed to a set of classifiers that assign attributes to each candidate. In parallel, each pair of candidate objects is processed by prepositional relationship functions which provide spatial relationships. A Conditional Random Field (CRF) is constructed that incorporates the unary image potentials computed in the previous steps and high order text based potentials computed from large document corpora. Finally, a labeling of the graph is predicted and a sentence is generated according to it. Other works~\cite{zitnick2013learning,hodosh2013framing, socher2010connecting} have focused on developing bi-directional mappings. Farhadi et al.~\cite{farhadi2010every} developed a method that can compute a score linking an image to a sentence. By using this score, a descriptive sentence from a fixed set can be given to a query image, or a relevant image from a fixed set can be found for a given query sentence. They suggested an intermediate space into which both the images and the sentences are mapped. In~\cite{karpathy2014deep}, Karpathy et al. introduce a model of bidirectional retrieval of images and sentences. Unlike previous works, they do not map images or sentences into a common space. Instead, their model works at a finer scale and embeds fragments of images and fragments of sentences into a common space. The sentence fragments are represented as dependency tree relations that are based on the dependence tree~\cite{socher2013grounded} of the sentence. The image fragments are represented by using a CNN~\cite{lecun1998gradient}. First, objects in the image are detected using Region Convolutional Neural Network (RCNN)~\cite{girshick2013rich}. The top 19 detected locations and the entire image are used as image fragments. Each image fragment is embedded using a CNN which takes the image inside a given bounding box and returns the embedding. Finally, they suggest a similarity score for any image-sentence pair. Their method achieved state-of-the-art results on image-sentence retrieval tasks on Pascal1K~\cite{rashtchian2010collecting}, Flickr8K~\cite{hodosh2013framing} and Flickr30K~\cite{hodoshimage} datasets. In our method, we return to the paradigm in which the images and sentences are mapped into a common domain and show significant improvement over the state-of-the-art for these three datasets. Similar to the previous work, we are using a CNN that takes an image as input and embeds it into a single vector by taking the representation of the last layer. The sentences are treated in a different way than in the previous work. Specifically, we employ word2vec~\cite{mikolov2013distributed} and map every word in the sentence to a vector. All of the vectors that belong to a sentence are then pooled into a single vector by using Fisher Vector with LMM and HGLMM distributions. Finally, the representations of the images and sentences are mapped into a common space by using the CCA algorithm~\cite{hotelling1936relations}.

Concurrently with our work, the field of image annotation has attracted a lot of attention. These very recent technical reports are listed below. In our experiments we compare directly with these methods.

Karpathy and Fei Fei~\cite{Karpathy} present a model that generates free-form natural language descriptions of image regions. Image caption  datasets are used to  developed a deep neural network model that infers the latent alignment between segments of sentences and the region of the image that is being describe. Their model associates the two modalities through a Markov Random Field (MRF) formulation. In addition, they introduced a multimodal Recurrent Neural Network (RNN) architecture that takes an input image and generates its description in text. 

Kiros et al.~\cite{kiros2014unifying} introduce an encoder-decoder pipeline that learns a multimodal joint embedding space with images and text, and a novel language model for decoding distributed representations from the multimodal space. This pipeline unifies joint image-text embedding models with multimodal neural language models. They introduce the structure-content neural language model that disentangles the structure of a sentence to its content, conditioned on representations produced by the encoder. The encoder allows one to rank images and sentences while the decoder can generate novel descriptions from scratch. They use long short-term memory (LSTM) to encode sentences, and the VGG~\cite{oxford} deep convolution neural network (CNN) to represent images.

Vinyals et al.~\cite{vinyals2014show} also describe a method of image description generation. Their work was inspired by recent advances in machine translation, where the task is to transform a sentence S written in a source language, into its translation T in the target language. Recent work approached this using RNN. An “encoder” RNN reads the source sentence and transforms it into a rich fixed-length vector representation, which in turn is used as the initial hidden state of a “decoder” RNN that generates the target sentence. Vinyals et al. suggested to use same recipe, replacing the encoder RNN by the GoogLeNet~\cite{DBLP:journals/corr/SzegedyLJSRAEVR14} CNN. 

Mao et al.~\cite{mao2014explain} presented a multimodal Recurrent Neural Network (m-RNN) model for generating novel sentence descriptions in order to explain the content of images. It directly models the probability distribution of generating a word given previous words and the image. Image descriptions are generated by sampling from this distribution. The model consists of two sub-networks: a deep recurrent neural network for sentences and a deep convolutional network for images. These two sub-networks interact with each other in a multimodal layer, creating the complete m-RNN model.

Donahue et al.~\cite{donahue} suggest the idea of convolutional networks which are also recurrent, or “temporally deep”. The authors develop a novel recurrent convolutional architecture suitable for large-scale visual learning, which is end-to-end trainable, and demonstrate the value of these models on benchmark video recognition tasks, image description and retrieval problems, and video narration challenges. In contrast to previous models that assume a fixed spatio-temporal receptive field or simple temporal averaging for sequential processing, recurrent convolutional models are “doubly deep” in that they can be compositional in spatial and temporal “layers”. Such models may have advantages when target concepts are complex or when the training data is limited.

\paragraph{Representing text as vectors}

Word2vec~\cite{word2vec} is a recently developed technique for building a neural network that maps words to real-number vectors, with the desideratum that words with similar meanings will map to similar vectors. This technique belongs to the class of methods called ``neural language models''. Using a scheme that is much simpler than previous work in this domain, where neural networks with many hidden units and several non-linear layers were normally constructed (e.g.,~\cite{Bengio:2003:NPL:944919.944966}), word2vec~\cite{word2vec} constructs a simple log-linear classification network~\cite{Mnih:2007:TNG:1273496.1273577}. Two such networks are proposed: the Skip-gram architecture and the Continuous Bag-of-words (CBOW) architecture. In our experiments, we employ the Skip-gram architecture, which is considered preferable.

Recently, the attention has shifted into representing sentences and paragraphs and not just words. The classical method in this domain is Bag of Words~\cite{BOW}. Socher et al.~\cite{SocherEtAl2011:RNN} have analyzed sentences using a recursive parse tree. A sentence is then represented by a matrix. In a recent contribution~\cite{ DBLP:conf/icml/LeM14} the neural network learns to predict the following word in a paragraph based on a representation that concatenates the vector representation of the previous text and the vector representations of a few words from the paragraph.


\section{Laplacian Mixture Model (LMM)}

The Laplacian Mixture Model (LMM) is a parametric probability density function represented as a weighted sum of multivariate Laplacians. The multivariate Laplacian itself represents a distribution over vectors in $R^{D}$. Note that unlike the multivariate Gaussian, it is not uniquely defined~\cite{eltoft2006multivariate}. In our formulation, similar to the underlying GMM distributions of conventional Fisher Vectors~\cite{perronnin2007fisher}, it is assumed that each multivariate Laplacian has a diagonal covariance matrix. Under this assumption, the probability density function of a single multivariate Laplacian is:

$$f(x;m_1,s_1) = \prod_{d=1}^{D}{\frac{1}{2s_{1,d}} exp\left(-\frac{|x_{d}-m_{1,d}|}{s_{1,d}}\right)}~,$$
where $m_1 \in R^{D}$ and $s_1 \in R^{D}$ are called the location parameter vector and the scale parameter vector of the multivariate Laplacian, and the second index $d$ is the index of the vector coordinates. The LMM is defined by a set of parameters $\lambda=\{\tau_{k},m_{k},s_{k}\}_{k=1 \dots K}$ where $\tau_{k} \in R$, $m_{k} \in R^{D}$, and $s_{k} \in R^{D}$ denote respectively the weight, location parameter vector and the scale parameter vector of the $k^{th}$ component and where K is the number of components in the mixture.

\subsection{Fitting a LMM}

Given a set of data points $\{x_{1},x_{2}, \dots , x_{N}\}$ one would like to estimate the parameters $\lambda$ of a LMM. Similar to a GMM, an expectation maximization~\cite{em} approach could be taken here. Specifically:

\paragraph{Expectation Step}

Let $Z_{i}=k$ be the event of $x_{i}$ being associated with mixture component $k$. Let $T_{k,i}^{(t)}=P\left(Z_{i}=k|X=x_{i};\lambda^{(t)}\right)$ be the conditional probability of sample $x_{i}$ being associated with component k, given the sample $x_{i}$ and the current estimation at iteration $t$ of the parameters $\lambda^{(t)}$. It is straightforward to show that:
$$T_{k,i}^{(t)}=\frac{\tau_{k}^{(t)} \cdot f(x_{i};m_{k}^{(t)},s_{k}^{(t)})}{\sum_{r=1}^{K}{\tau_{r}^{(t+1)} \cdot f(x_{i};m_{r}^{(t)},s_{r}^{(t)})}}$$
where $f(x_{i};m_{k}^{(t)},s_{k}^{(t)})$ is the pdf of the $k^{th}$ multivariate Laplacian evaluated at point $x_{i}$ according to the current estimation of parameters, $\lambda^{(t)}$.

\paragraph{Maximization Step}

Let $Q\left(\lambda | \lambda^{(t)}\right)$ be the expected value of the log likelihood function, with respect to the conditional distribution of $Z$ given $X$ under the current estimate of the parameters $\lambda^{(t)}$.

$$Q\left(\lambda | \lambda^{(t)}\right)=E_{Z|X,\lambda^{(t)}}\left[log\left(L\left(\lambda;X,Z\right)\right)\right]~~.$$

By deriving $Q\left(\lambda | \lambda^{(t)}\right)$ according to $\lambda$ and solving the resulting equations (the full details are given in the supplementary material), one gets the following maximization step expressions:

\begin{equation}\label{eq:lmm_maxstep_tau}\tau_{k}^{(t+1)} = \frac{\sum_{i=1}^{N}T_{k,i}^{\left(t\right)}}{\sum_{r=1}^{K}\sum_{i=1}^{N}T_{r,i}^{\left(t\right)}}\end{equation}

\begin{equation}\label{eq:lmm_maxstep_m}\sum_{m_{k,d}^{(t+1)} \leq x_{i,d}} T_{k,i}^{\left(t\right)} = \sum_{m_{k,d}^{(t+1)} > x_{i,d}} T_{k,i}^{\left(t\right)}\end{equation}

\begin{equation}\label{eq:lmm_maxstep_s}s^{(t+1)}_{k,d}=\frac{\sum_{i=1}^{N}T_{k,i}^{\left(t\right)}\left|x_{i,d}-m_{k,d}^{(t+1)}\right|}{\sum_{i=1}^{N}T_{k,i}^{\left(t\right)}}\end{equation}

Unlike the EM for GMMs, where the mean is provided explicitly, the location parameter for LMMs is not explicitly given. Rather it is defined as the weighted median of equation~\ref{eq:lmm_maxstep_m}. An efficient solution to equation~\ref{eq:lmm_maxstep_m} would be to sort the values of $x_{i,d}$ for all the samples $i$ and then to iterate over the sorted values and to choose a value for $m_{k,d}^{(t+1)}$ which minimizes the gap between the expressions at both sides of the equation $\left| \sum_{m_{k,d}^{(t+1)} \leq x_{i,d}} T_{k,i}^{\left(t\right)} - \sum_{m_{k,d}^{(t+1)} > x_{i,d}} T_{k,i}^{\left(t\right)}\right|$.

\subsection{Fisher Vector of a LMM}

The Fisher Vector that was introduced in \cite{perronnin2007fisher} are the gradients of the log-likelihood of the data with respect to the parameters of the GMM. By following the same path, a variant of the Fisher Vector can be derived for the LMM. Let $X=\{x_{1},x_{2}, \dots , x_{N}\}$ be a set of samples and $\lambda$ are the parameters of a LMM. Denote the log-likelihood of the samples $X$ with respect to the parameters $\lambda$ by $\mathcal{L}\left(X|\lambda\right)$. Then the Fisher Vector of the LMM is (see supplementary material):

\begin{equation}\label{eq:lmm_fisher_m}\frac{\partial{\mathcal{L}\left(X|\lambda\right)}}{\partial{m_{k,d}}}=\sum_{i=1}^{N} \frac{T_{k,i}}{s_{k,d}} \cdot \left\{\begin{array}{l l}
1 & \quad \text{if $x_{i,d}>m_{k,d}$}\\
-1 & \quad \text{otherwise}
\end{array}\right.\end{equation}

\begin{equation}\label{eq:lmm_fisher_s}\frac{\partial{\mathcal{L}\left(X|\lambda\right)}}{\partial{s_{k,d}}}=\sum_{i=1}^{N} T_{k,i} \left(\frac{\left|x_{i,d}-m_{k,d}\right|}{s_{k,d}^2} - \frac{1}{s_{k,d}}\right)\end{equation}

As in \cite{perronnin2007fisher}, the diagonal of the Fisher Information Matrix $F$ is approximated in order to normalize the dynamic range of the different dimensions of the gradient vectors. Denote by $f_{m_{k,d}}$ and $f_{s_{k,d}}$ the terms of the diagonal of $F$ which correspond respectively to $\frac{\partial{\mathcal{L}\left(X|\lambda\right)}}{\partial{m_{k,d}}}$ and $\frac{\partial{\mathcal{L}\left(X|\lambda\right)}}{\partial{s_{k,d}}}$. Therefore, the normalized partial derivatives are $f_{m_{k,d}}^{-1/2}\cdot\frac{\partial{\mathcal{L}\left(X|\lambda\right)}}{\partial{m_{k,d}}}$ and $f_{s_{k,d}}^{-1/2}\cdot\frac{\partial{\mathcal{L}\left(X|\lambda\right)}}{\partial{s_{k,d}}}$. It is shown in the supplementary that $f_{m_{k,d}}$ and $f_{s_{k,d}}$ are approximately:

$$f_{m_{k,d}}=\frac{N\tau_{k}}{s_{k,d}^{2}}$$
$$f_{s_{k,d}}=\frac{N\tau_{k}}{s_{k,d}^{2}}$$

\subsection{LMM and ICA}

In \cite{perronninFVTheoryPractice}, Sanchez et al. state that applying the Principal Components Analysis (PCA) on the data before fitting the GMM is key to make the Fisher Vector work. In experiments on PASCAL VOC 2007, they show that accuracy does not seem to be overly sensitive to the exact number of PCA components. The explanation is that transforming the descriptors by using PCA is a better fit to the diagonal covariance matrix assumption.

Following this observation, a transformation that will cause the transformed descriptors to be a better fit to the diagonal covariance matrix assumption is sought for the LMM. The optimal transformation will result with transformed descriptors that are dimension independent and are non-Gaussian signals. Therefore, and since PCA suffers from the implicit assumption of underlying Gaussian Distribution~\cite{ke2004pca}, the Independent Component Analysis (ICA)~\cite{journals/tnn/Lee99a} is chosen in our experiments. It seems that, for the image-text matching experiments we run, ICA is not only preferable when using LMM, but also preferable when using GMM.

\section{Hybrid Gaussian-Laplacian Mixture Model}

By combining the Gaussian and the Laplacian distributions into one hybrid distribution model one can hope to benefit from the properties of the two distributions. We define the Hybrid Gaussian-Laplacian unnormalized distribution, $h(x;\mu,\sigma,m,s,b)$, for the univariate case to be:
$$h(x;\mu,\sigma,m,s,b)=l(x;m,s)^{b} \cdot g(x;\mu,\sigma)^{1-b}~,$$
where $l(x;m,s)$ is the Laplacian distribution parameterized with location parameter $m$ and scale parameter $s$ and $g(x;\mu,\sigma)$ is the Gaussian distribution parameterized by the mean $\mu$ and the standard deviation $\sigma$. The parameter $b$ is constrained to be in the range $\left[0,1\right]$. According to these definitions, the Hybrid Gaussian-Laplacian Mixture Model is a weighted geometric mean of the Laplacian distribution and the Gaussian distribution. Geometric means of distributions are used, for example, to perform smooth transition between distributions in a simulated annealing framework~\cite{radford2001}. More recently, geometric means have emerged as the distribution of the averaged predictions of multiple neural networks employing a softmax layer~\cite{DBLP:journals/corr/abs-1207-0580}.

As before in the case of LMM, the Hybrid Gaussian-Laplacian can be defined for the multivariate case by assuming that the dimensions are independent. Under this assumption, the probability density function of a single multivariate Hybrid Gaussian-Laplacian is:

$$\prod_{d=1}^{D}{h(x_{d};\mu_{d},\sigma_{d},m_{d},s_{d},b_{d})}$$

The HGLMM is a mixture model of the Hybrid Gaussian-Laplacian multivariate distribution. It is defined by a set of parameters $\lambda=\{\tau_{k},\mu_{k},\sigma_{k},m_{k},s_{k},b_{k}\}_{k=1 \dots K}$, where $\tau_{k} \in R$ is the weight of the $k^{th}$ component and $\mu_{k},\sigma_{k},m_{k},s_{k},b_{k} \in R^{D}$ are the parameters of the Gaussian and Laplacian distributions and the weights of the geometric mean.

\subsection{Fitting a HGLMM}

Once again, the EM algorithm can be used in order to estimate the parameters $\lambda$ of the HGLMM given a set of data points $\{x_{1},x_{2}, \dots , x_{N}\}$.

\paragraph{Expectation Step}

Let $T_{k,i}^{(t)}=P\left(Z_{i}=k|X=x_{i};\lambda^{(t)}\right)$ then for the HGLMM the following equation holds:

$$T_{k,i}^{(t)}=\frac{\tau_{k}^{(t)} \cdot h(x;\mu_{k}^{t},\sigma_{k}^{t},m_{k}^{t},s_{k}^{t},b_{k}^{t})}{\sum_{r=1}^{K}{\tau_{r}^{(t)} \cdot h(x;\mu_{r}^{t},\sigma_{r}^{t},m_{r}^{t},s_{r}^{t},b_{r}^{t})}}$$

Note that, in essence, we sum multiple unnormalized distributions. However, as shown below, in practice the hybrid distribution $h$ is a normalized one.

\paragraph{Maximization Step}

As before, the expected value of the log likelihood function, with respect to the conditional distribution of $Z$ given $X$ under then current estimate of the parameters $\lambda^{(t)}$ is computed:

$$Q\left(\lambda | \lambda^{(t)}\right)=E_{Z|X,\lambda^{(t)}}\left[log\left(L\left(\lambda;X,Z\right)\right)\right]$$.

Deriving $Q\left(\lambda | \lambda^{(t)}\right)$ according to the parameters $\{\tau,\mu,\sigma, m ,s \}$ and solving the resulting equations
yields the same expressions that one would get for the maximization step in the GMM and in the LMM, up to the values of $T_{k,i}^{t}$ which are different and are defined in the Expectation Step. Specifically, the equations for $\tau_{k}$, $m_k$, and $s_k$ are given above in equations~\ref{eq:lmm_maxstep_tau},~\ref{eq:lmm_maxstep_m},~\ref{eq:lmm_maxstep_s}, and the equations for $\mu_k$ and $\sigma_k$ are as follows:

\begin{equation}\mu_{k,d}^{(t+1)}=\frac{\sum_{i=1}^{N}T_{k,i}^{\left(t\right)}\cdot x_{i,d}}{\sum_{i=1}^{N}T_{k,i}^{\left(t\right)}}\end{equation}

\begin{equation}(\sigma_{k,d}^{(t+1)})^{2}=\frac{\sum_{i=1}^{N}T_{k,i}^{\left(t\right)}\left(x_{i,d}-\mu_{k,d}^{(t+1)}\right)^{2}}{\sum_{i=1}^{N}T_{k,i}^{\left(t\right)}}\end{equation}


In order to maximize the log-likelihood of the data according to the parameters $b$ one can simply look at the contribution of the parameter $b_{k,d}$ to the log-likelihood. Specifically, omitting the iteration index, let:

$$L_{b_{k,d}}=\sum_{i=1}^{N}T_{k,i}^{\left(t\right)}\left(-\log\left(2s_{k,d}\right)-\frac{|x_{i,d}-m_{k,d}|}{s_{k,d}}\right)$$
$$G_{b_{k,d}}=\sum_{i=1}^{N}T_{k,i}^{\left(t\right)}\left(-\log\left(\sqrt{2\pi}\sigma_{k,d}\right)-\frac{\left(x_{i,d}-\mu_{k,d}\right)^{2}}{2\sigma_{k,d}^{2}}\right)$$

Then the contribution of $b_{k,d}$ to the log-likelihood is:

\begin{equation}b_{k,d} \cdot L_{b_{k,d}} + \left(1-b_{k,d}\right) \cdot G_{b_{k,d}}\end{equation}

Therefore, under the constraint that $0\leq b_{k,d} \leq 1$, the value of $b_{k,d}^{(t+1)}$ that maximizes the log-likelihood is:

\begin{equation}\label{eq:hglmm_maxstep_b}b_{k,d}^{(t+1)}=\left\{\begin{array}{l l}
1 & \quad \text{if $L_{b_{k,d}}^{(t+1)}>G_{b_{k,d}}^{(t+1)}$}\\
0 & \quad \text{otherwise}
\end{array}\right.\\ \end{equation}

Due to equation~\ref{eq:hglmm_maxstep_b}, after the maximization step, the distribution of each dimension in each component of the HGLMM is either a Gaussian or Laplacian. Therefore, the final output of the EM algorithm also contains sharp selections between the two distributions, and the hybrid mixture model is a normalized probability model.

\subsection{Fisher Vector of a HGLMM}

A Fisher Vector variant can also be defined for the HGLMM. Although the HGLMM has more parameters than the GMM, the Fisher Vector of the HGLMM has the same length as the Fisher Vector of the GMM which is $2KD$. The reason is that according to equation~\ref{eq:hglmm_maxstep_b}, each dimension $d$ in each component $k$ is either a Laplacian univariate distribution or a Gaussian univariate distribution and therefore the contribution to the Fisher Vector is $\left\{\frac{\partial\mathcal{L}(X|\lambda)}{\partial\mu_{k,d}},\frac{\partial\mathcal{L}(X|\lambda)}{\partial\sigma_{k,d}}\right\}$ if $b_{k,d}=0$ and $\left\{\frac{\partial\mathcal{L}(X|\lambda)}{\partial m_{k,d}},\frac{\partial\mathcal{L}(X|\lambda)}{\partial s_{k,d}}\right\}$ if $b_{k,d}=1$. The values of the Fisher Vector for the HGLMM are computed by taking the gradients of the log-likelihood of the data with respect to the parameters of the HGLMM. The resulting equations yields the same expressions that one would get for the Fisher Vector of the GMM and of the LMM, up to the values of $T_{k,i}$ which are different and have the same expressions as in the expectation step of the HGLMM. Therefore the values of the Fisher Vector for the HGLMM when $b_{k,d}=0$ are:

\begin{equation}\label{eq:hglmm_fisher_mu}\frac{\partial{\mathcal{L}\left(X|\lambda\right)}}{\partial{\mu_{k,d}}}=\sum_{i=1}^{N} T_{k,i} \cdot \frac{x_{i,d} - \mu_{k,d}}{\sigma_{k,d}^2}\end{equation}

\begin{equation}\label{eq:hglmm_fisher_sigma}\frac{\partial{\mathcal{L}\left(X|\lambda\right)}}{\partial{\sigma_{k,d}}}=\sum_{i=1}^{N} T_{k,i} \left(\frac{\left(x_{i,d}-\mu_{k,d}\right)^{2}}{\sigma_{k,d}^3} - \frac{1}{\sigma_{j,d}}\right)\end{equation}

When $b_{k,d}=1$, equations~\ref{eq:lmm_fisher_m} and~\ref{eq:lmm_fisher_s} provide the relevant Fisher Vector coordinates.

%
%

As in the LMM, the diagonal of the Fisher Information Matrix F is approximated in order to normalize the dynamic range of the different dimensions of the gradient vectors. Let $f_{\mu_{k,d}}$, $f_{\sigma_{k,d}}$, $f_{m_{k,d}}$ and $f_{s_{k,d}}$ be the terms of the diagonal of $F$ that correspond respectively to $\frac{\partial{\mathcal{L}\left(X|\lambda\right)}}{\partial{\mu_{k,d}}}$, $\frac{\partial{\mathcal{L}\left(X|\lambda\right)}}{\partial{\sigma_{k,d}}}$, $\frac{\partial{\mathcal{L}\left(X|\lambda\right)}}{\partial{m_{k,d}}}$ and $\frac{\partial{\mathcal{L}\left(X|\lambda\right)}}{\partial{s_{k,d}}}$. Then, the normalized partial derivatives are
$f_{m_{k,d}}^{-1/2}\cdot\frac{\partial{\mathcal{L}\left(X|\lambda\right)}}{\partial{\mu_{k,d}}}$, $f_{s_{k,d}}^{-1/2}\cdot\frac{\partial{\mathcal{L}\left(X|\lambda\right)}}{\partial{\sigma_{k,d}}}$,
$f_{m_{k,d}}^{-1/2}\cdot\frac{\partial{\mathcal{L}\left(X|\lambda\right)}}{\partial{m_{k,d}}}$ and $f_{s_{k,d}}^{-1/2}\cdot\frac{\partial{\mathcal{L}\left(X|\lambda\right)}}{\partial{s_{k,d}}}$. It can be shown that the approximated values of the terms of the diagonal are the same as the approximated values of the terms of the diagonal in the GMM and LMM:

$$f_{\mu_{k,d}}=\frac{N\tau_{k}}{\sigma_{k,d}^{2}},f_{\sigma_{k,d}}=\frac{2N\tau_{k}}{\sigma_{k,d}^{2}},f_{m_{k,d}}=\frac{N\tau_{k}}{s_{k,d}^{2}},f_{s_{k,d}}=\frac{N\tau_{k}}{s_{k,d}^{2}}$$

\begin{table*}
\centering
\begin{tabular}{|l|lllll|lllll|l|}
\hline
& \multicolumn{5}{|c|}{Image search} & \multicolumn{5}{|c|}{Image annotation} & {Sentence}\\
& r@1 & r@5 & r@10 & median & mean & r@1 & r@5 & r@10 & median & mean & mean \\
& & & & rank & rank & & & & rank & rank& rank\\
\hline
GCS~\cite{socher2013grounded} & 6.1 & 18.5 & 29.0 & 29.0 & NA & 4.5 & 18.0 & 28.6 & 32.0 & NA & NA \\
SDT-RNN~\cite{socher2013grounded} & 6.6 & 21.6 & 31.7 & 25.0 & NA & 6.0 & 22.7 & 34.0 & 23.0 & NA & NA \\
DFE~\cite{karpathy2014deep} & 9.7 & 29.6 & 42.5 & 15.0 & NA & 12.6 & 32.9 & 44.0 & 14.0 & NA & NA \\
BRNN~\cite{Karpathy}   & 11.8 & 32.1 & 44.7 & 12.4 & NA & 16.5 & 40.6 & 54.2 & 7.6  & NA & NA \\
SC-NLM~\cite{kiros2014unifying}  & 12.5 & 37.0 & 51.5 & 10.0 & NA & 18.0 & 40.9 & 55.0 & 8.0  & NA & NA \\
NIC~\cite{vinyals2014show}  & 19.0 & NA   & 64.0 & 5.0  & NA & 20.0 & NA   & 61.0 & 6.0  & NA & NA \\
m-RNN~\cite{mao2014explain} & 11.5 & 31.0 & 42.4 & 15.0 & NA & 14.5 & 37.2 & 48.5 & 11.0 & NA & NA \\
\hline
Overfeat~\cite{sermanet2013overfeat}: & & & & & & & & & & &\\
~Mean Vec & 11.3 & 30.9 & 44.0 & 14.0 & 48.0 & 14.3 & 33.9 & 45.3 & 13.0 & 59.0 & 13.6 \\
~GMM & 12.3 & 33.4 & 46.0 & 13.0 & 41.5 & 18.0 & 40.0 & 50.1 & 10.0 & 46.1 & 12.7 \\
~LMM & 12.6 & 32.9 & 46.0 & 13.0 & 42.8 & 18.9 & 39.2 & 49.9 & 11.0 & 46.8 & 13.6 \\
~HGLMM & 12.7 & 33.6 & 46.6 & 12.0 & 42.0 & 19.4 & 40.5 & 50.9 & 9.0 & 46.0 & 12.9 \\
~GMM+HGLMM & 13.0 & 34.2 & 47.3 & 12.0 & 40.3 & 17.9 & 41.2 & 51.5 & 10.0 & 43.8 & 12.3\\
\hline
VGG~\cite{oxford}: & & & & & & & & & & & \\
~Mean Vec  & 19.1 & 45.3 & 60.4 & 7.0 & 27.0 & 22.6 & 48.8 & 61.2 & 6.0 & 28.7 & 12.5 \\
~GMM   & 20.6  & 48.6 & 64.0 & 6.0 & 31.6 & 28.4 & 57.7 & 70.0   & 4.0 & 24.7 & 12.0 \\
~LMM   & 19.8 & 47.6 & 62.6 & 6.0 & 33.7 & 27.7 & 56.5 & 69.0   & 4.0 & 25.5 & 13.0 \\
~HGLMM & 20.2 & 48.0 & 62.7 & 6.0 & 30.8 & 29.3 & 58.1 & 71.9 & 4.0 & 22.1 & 14.0 \\
~GMM+HGLMM   & 21.3 & 50.0 & 64.6 & 5.0 & 30.3 & 31.0   & 59.3 & 73.7 & 4.0 & 22.2 & 11.9 \\
\hline
\end{tabular}
\vspace{.2cm}
\caption{Results on the Flickr8K benchmark~\cite{hodosh2013framing}. Shown are the recall rates at $1,5$ and $10$ retrieval results (higher is better). Also shown, the mean and median rank of the first ground truth (lower is better).
There are three tasks: image annotation, image search, and sentence similarity.
We compare the results of \cite{socher2013grounded,karpathy2014deep,Karpathy,kiros2014unifying,vinyals2014show,vinyals2014show,mao2014explain} to the mean vector baseline and to Fisher Vectors based on GMM, LMM and HGLMM. In addition we report results for the combination of the GMM and HGLMM Fisher Vectors.
}
\label{tab:f8k}
\end{table*} 

\begin{table*}
\centering
\begin{tabular}{|l|lllll|lllll|l|}
\hline
& \multicolumn{5}{|c|}{Image search} & \multicolumn{5}{|c|}{Image annotation} & {Sentence}\\
& r@1 & r@5 & r@10 & median & mean & r@1 & r@5 & r@10 & median & mean & mean \\
& & & & rank & rank & & & & rank & rank& rank\\
\hline
DFE~\cite{karpathy2014deep} & 10.3 & 31.4 & 44.5 & 13.0 & NA    & 16.4 & 40.2 & 54.7 & 8.0  & NA    & NA    \\
BRNN~\cite{Karpathy}                & 15.2  & 37.7  & 50.5  & 9.2 & NA    & 22.2 & 48.2 & 61.4 & 4.8 & NA    & NA    \\
SC-NLM~\cite{kiros2014unifying}     & 16.8  & 42.0    & 56.5  & 8.0   & NA    & 23.0   & 50.7 & 62.9 & 5.0   & NA    & NA    \\
NIC~\cite{vinyals2014show}         & 17.0    & NA    & 57.0    & 7 .0  & NA    & 17.0   & NA   & 56.0   & 7.0   & NA    & NA    \\
m-RNN~\cite{mao2014explain}    & 12.6  & 31.2  & 41.5  & 16.0  & NA    & 18.4 & 40.2 & 50.9 & 10.0  & NA    & NA    \\
LRCN~\cite{donahue}    & 14.0    & 34.9  & 47.0    & 11.0  & NA    & NA   & NA   & NA   & NA  & NA    & NA    \\
\hline
Overfeat~\cite{sermanet2013overfeat}: & & & & & & & & & & &\\
Mean Vec & 11.1 & 29.3 & 41.9 & 16.0 & 63.8 & 14.3 & 34.0 & 43.9 & 15.0 & 73.8 & 16.9 \\
GMM & 14.2 & 34.9 & 47.7 & 12.0 & 50.0 & 20.8 & 42.7 & 53.2 & 9.0  & 61.9 & 17.6 \\
LMM & 14.0 & 34.7 & 47.3 & 12.0 & 53.2 & 20.0 & 41.0 & 53.1 & 9.0  & 63.0 & 18.4 \\
HGLMM & 14.9 & 36.0 & 48.3 & 11.0 & 50.8 & 20.9 & 41.7 & 53.7 & 8.0  & 57.9 & 16.5 \\
GMM+HGLMM & 14.6 & 36.6 & 49.1 & 11.0 & 48.9 & 21.2 & 41.2 & 55.4 & 8.0  & 56.2 & 15.9 \\
\hline
VGG~\cite{oxford}: & & & & & & & & & & & \\
~Mean Vec  & 20.7 & 47.2 & 60.6 & 6.0 & 29.9 & 24.5 & 53.4 & 65.6 & 5.0 & 24.9 & 14.9 \\
~GMM       & 24.2 & 51.7 & 65.2 & 5.0 & 33.7 & 31.8 & 59.3 & 71.7 & 3.0 & 20.0 & 17.0 \\
~LMM       & 23.8 & 51.7 & 65.4 & 5.0 & 33.0 & 31.2 & 60.6 & 73.1 & 3.0 & 21.1 & 15.6 \\
~HGLMM     & 24.9 & 52.3 & 66.4 & 5.0 & 32.5 & 33.2 & 60.7 & 72.4 & 3.0 & 18.8 & 15.8 \\
~GMM+HGLMM & 25.6 & 53.2 & 66.8 & 5.0 & 31.3 & 33.3 & 62.0 & 74.7 & 3.0 & 18.7 & 15.6\\
\hline
\end{tabular}
\vspace{.2cm}
\caption{Mean results on the Flickr30K benchmark~\cite{hodoshimage}. For details, see Table~\ref{tab:f8k}. GCS and SDT-RNN results~\cite{socher2013grounded} are not available for this specific benchmark.}
\label{tab:f30k}
\end{table*}

\begin{table*}
\centering
\begin{tabular}{|l|lllll|lllll|l|}
\hline
& \multicolumn{5}{|c|}{Image search} & \multicolumn{5}{|c|}{Image annotation} & {Sentence}\\
& r@1 & r@5 & r@10 & median & mean & r@1 & r@5 & r@10 & median & mean & mean \\
& & & & rank & rank & & & & rank & rank& rank\\
\hline
GCS~\cite{socher2013grounded} & 16.4 & 46.6 & 65.6 & NA  & 12.5 & 23.0 & 45.0 & 63.0 & NA  & 16.9 & 10.5 \\
SDT-RNN~\cite{socher2013grounded} & 25.4 & 65.2 & 84.4 & NA  & 7.0  & 25.0 & 56.0 & 70.0 & NA  & 13.4 & NA   \\
DFE~\cite{karpathy2014deep} & 23.6 & 65.2 & 79.8 & NA  & 7.6  & 39.0 & 68.0 & 79.0 & NA  & 10.5 & NA   \\
\hline
Overfeat~\cite{sermanet2013overfeat}: & & & & & & & & & & &\\
~Mean Vec  & 34.2 & 75.7 & 88.6 & 2.1 & 5.2 & 39.8 & 71.8 & 84.0 & 2.3 & 6.9 & 2.1 \\
~GMM       & 33.2 & 73.9 & 84.9 & 2.4 & 7.6 & 42.6 & 73.7 & 83.9 & 2.1 & 8.5 & 2.4 \\
~LMM       & 34.5 & 74.9 & 85.7 & 2.2 & 7.2 & 42.7 & 74.5 & 84.8 & 1.9 & 8.6 & 2.3 \\
~HGLMM     & 33.7 & 74.5 & 85.6 & 2.3 & 7.4 & 41.8 & 73.1 & 84.5 & 2.3 & 8.5 & 2.5 \\
~GMM+HGLMM & 34.6 & 75.2 & 86.1 & 2.2 & 7.1 & 43.3 & 74.3 & 85.1 & 2.0 & 8.4 & 2.3\\
\hline
VGG~\cite{oxford}: & & & & & & & & & & & \\
~Mean Vec  & 44.9 & 84.9 & 94.2 & 2.0 & 3.5 & 52.5 & 83.3 & 92.3 & 1.3 & 4.0 & 2.0 \\
~GMM       & 43.2 & 84.7 & 92.7 & 2.0 & 4.5 & 55.4 & 85.7 & 92.8 & 1.1 & 4.3 & 2.3 \\
~LMM       & 42.9 & 83.4 & 92.2 & 2.0 & 4.6 & 54.7 & 84.5 & 93.0 & 1.1 & 4.1 & 2.4 \\
~HGLMM     & 43.2 & 84.0 & 92.4 & 2.0 & 4.7 & 55.4 & 85.4 & 93.1 & 1.2 & 4.3 & 2.4 \\
~GMM+HGLMM & 43.9 & 84.8 & 93.2 & 2.0 & 4.4 & 56.3 & 85.9 & 93.2 & 1.2 & 4.1 & 2.3 \\
\hline
\end{tabular}
\vspace{.2cm}
\caption{Mean results on the Pascal1K benchmark~\cite{rashtchian2010collecting}. For details, see Table~\ref{tab:f8k}. This dataset is rather small and the contribution of high dimensional  representations is less obvious than in the Flickr datasets.}
\label{tab:p1k}
\end{table*}

\begin{figure}[t]
\centering
\includegraphics[width=1.0\linewidth]{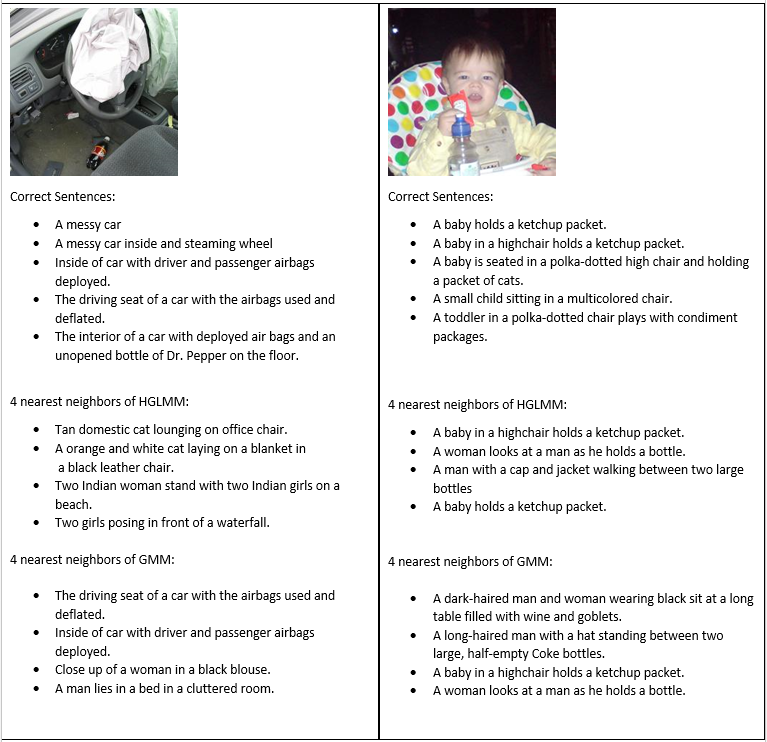}
\caption{\small Shown are two examples in which GMM Fisher Vectors and HGLMM Fisher Vectors considerably differ. For the left example, GMM's rank one result was correct. The rank of the first ground truth result in the list of HGLMM was 35. In the example on the right, the corresponding ranks were 3 and 1.}
\label{fig:sequences}
\end{figure}

\begin{table*}
\centering
\begin{tabular}{|l|lllll|lllll|l|}
\hline
& \multicolumn{5}{|c|}{Image search} & \multicolumn{5}{|c|}{Image annotation} & {Sentence}\\
& r@1 & r@5 & r@10 & median & mean & r@1 & r@5 & r@10 & median & mean & mean \\
& & & & rank & rank & & & & rank & rank& rank\\
\hline
1K test images:                                                                    &      &      &      &      &      &      &      &      &      &      &      \\
~BRNN~\cite{Karpathy}                                                                              & 20.9 & 52.8 & 69.2 & 4.0  & NA   & 29.4 & 62.0 & 75.9 & 2.5  & NA   & NA   \\
~Mean Vec                                                                      & 24.2 & 56.4 & 72.4 & 4.0  & 14.7 & 33.2 & 61.8 & 75.1 & 3.0  & 14.5 & 14.3 \\
~GMM                                                                 & 24.7 & 58.7 & 75.6 & 4.0  & 13.3 & 38.3 & 66.1 & 79.2 & 3.0  & 12.1 & 13.9 \\
~LMM                                                                 & 24.8 & 59.2 & 75.4 & 4.0  & 12.8 & 39.1 & 68.4 & 79.4 & 2.0  & 12.2 & 13.9 \\
~HGLMM                                                               & 25.1 & 59.7 & 76.5 & 4.0  & 12.7 & 38.7 & 68.4 & 81.0 & 2.0  & 11.3 & 13.7 \\
~GMM+HGLMM                                                                                & 25.6 & 60.4 & 76.8 & 4.0  & 12.3 & 38.9 & 68.4 & 80.1 & 2.0  & 10.5 & 13.2 \\
\hline
5K test images:                                                                    &      &      &      &      &      &      &      &      &      &      &      \\
~BRNN~\cite{Karpathy}                                                                              & 8.9  & 24.9 & 36.3 & 19.5 & NA   & 11.8 & 32.5 & 45.4 & 12.2 & NA   & NA   \\
~Mean Vec                                                                     & 10.3 & 27.2 & 38.4 & 18.0 & 64.7 & 12.8 & 32.1 & 44.6 & 14.0 & 62.2 & 63.7 \\
~GMM                                                                 & 10.5 & 28.0 & 39.4 & 17.0 & 62.3 & 17.0 & 38.1 & 49.8 & 11.0 & 52.2 & 58.0 \\
~LMM                                                                 & 10.5 & 28.0 & 39.7 & 17.0 & 61.3 & 16.4 & 38.1 & 50.4 & 10.0 & 51.2 & 57.0 \\
~HGLMM                                                               & 11.1 & 28.7 & 40.2 & 16.0 & 59.0 & 16.7 & 38.4 & 51.0 & 10.0 & 47.8 & 56.2 \\
~GMM+HGLMM & 11.2 & 29.2 & 41.0 & 16.0 & 57.2 & 17.7 & 40.1 & 51.9 & 10.0 & 45.5 & 54.5 \\
\hline
\end{tabular}
\vspace{.2cm}
\caption{Mean results on the COCO benchmark~\cite{coco}. For details, see Table~\ref{tab:f8k}. Results are shown only for our method where the visual representation is based on the VGG CNN~\cite{oxford}.}
\label{tab:coco}
\end{table*}

\section{Results}

We perform our image annotation experiments on four benchmarks: Pascal1K~\cite{rashtchian2010collecting}, Flickr8K~\cite{hodosh2013framing},  Flickr30K~\cite{hodoshimage}, and COCO~\cite{coco}. The datasets contain 1,000, 8,000, 30,000, and 123,000 images respectively. The annotation of the images was done using crowdsourcing via Amazon Mechanical Turk, with five independent sentences provided by five users to each image.

The Flickr8k dataset is provided with a training split of size 6091, a validation split of 1000 images, and a test split of size 1000. We use the same split. For Pascal1K, no training splits are given, and we use 20 splits of the same size used to report results in previous work: 800 train, 100 validation and 100 test images. The situation for Flickr30K is similar, and following previous work, we use 5 random splits of 1000 images for test, 1000 for validation, and the rest for training. For COCO, we follow Karpathy et al. ~\cite{Karpathy} and use 5000 images for both validation and testing, and also report results on a subset of 1000 testing images. 5 random splits were used here as well. The reduced number of repeats on Flickr30K and COCO compared to Pascal1K stems from the datasets size and the resulting computational burden.

The word2vec~\cite{mikolov2013distributed} vectors used to represent words were obtained from~\url{code.google.com/p/word2vec/}. We found that performing ICA without reducing the dimensionality of the 300D word2vec vectors helps performance in all methods, including the vanilla GMM based Fisher Vectors. The original vectors and PCA provided somewhat lower results. We therefore apply ICA to the word2vec representation in all of our experiments below. All images were resized to a fixed size of 221 by 221 pixels and encoded as a single vector using the Overfeat~\cite{sermanet2013overfeat} software. When employing the VGG~\cite{oxford} representation, the recommended pipeline is used. Namely, the original image is cropped in ten different ways into 224 by 224 pixel images: the four corners, the center, and their x-axis mirror image. The mean intensity is then subtracted in each color channel and the resulted images are encoded by the network. The average of the resulting 10 feature vectors is used as the single image representation.

Our system has two parameters: the number of components in the mixture models used to describe the sentence, and the regularization parameters of the regularized CCA algorithm~\cite{Vinod1976147} that matches between the image representation and the sentence representation. The first parameter was fixed to $30$ throughout the experiments. This value was selected once using the Flickr8K validation split. The regularization parameter was selected in each repetition of each experiment based on the validation data. We use linear CCA, which is presumably sub-optimal compared to the Kernel CCA that~\cite{socher2013grounded} used as one of the baseline methods.

There are three tasks: image annotation, in which the goal is to retrieve, given a query image, the five ground truth sentences; image search, in which, given a query sentence, the goal is to retrieve the ground truth image; and sentence similarity, in which the goal is to retrieve given a query sentence the other four sentences associated with the same image. For the first two tasks, the results are reported as the recall rate at one result, at 5 results, or at the 10 first results. Also reported is the mean rank and the median rank of the first ground truth result. In the sentence similarity task, only the mean rank of the first ground truth result is reported. 

In all tasks, CCA is trained on the training set, its parameter is tuned on the validation set, and testing is performed on the test set. For sentence similarity, the representations of the test samples are used after projection to the CCA space, however, the test images are not used. Note that all the parameters of the sentence representations, prior to the CCA computation, are learned in an unsupervised fashion on the corpus of word2vec vectors, without employing the three benchmark datasets.
 
Given a sentence, it is mapped to the set of word2vec vectors that are associated with the sentence's words. In our experiments, we test the average vector of this set as a simple baseline representation, three types of Fisher Vector pooling (GMM, LMM, and HGLMM), and fusion of GMM with HGLMM.  This fusion is done by concatenating the two Fisher Vector representations into a single vector. All Fisher Vectors were normalized by the L2 norm after applying the power normalization function with $\alpha=0.5$, as described in Sec.~\ref{sec:prev}.

We compare our results to the current state of the art, namely to the Grounded Compositional Semantics (GCS)~\cite{socher2013grounded}, Semantic Dependency Tree Recursive Neural Network (SDT-RNN)~\cite{socher2013grounded}, Deep Fragment Embedding Multi Instance Learning (DFE-MIL)~\cite{karpathy2014deep}, and the five concurrent technical reports described in Sec.~\ref{sec:prev}:  BRNN~\cite{Karpathy}, SC-NLM~\cite{kiros2014unifying}, NIC~\cite{vinyals2014show}, m-RNN~\cite{mao2014explain}, LRCN~\cite{donahue}.

The results are reported in Table ~\ref{tab:f8k},~\ref{tab:f30k},~\ref{tab:p1k},~\ref{tab:coco} for the Flickr8K, Flickr30K, Pascal1K, and COCO benchmarks respectively. As can be seen, the Fisher Vector methods outperform the existing state of the art and the mean vector baseline. As a general trend, in all three benchmarks, Fisher Vectors based on GMM outperform those based on LMM. HGLMM based Fisher Vectors perform better than both, and combining GMM with HGLMM outperforms the other methods. Sample results can be seen in Figure~\ref{fig:sequences}.

\subsection{Sentence Synthesis}

\begin{figure}
\centering
\includegraphics[width=1.0\linewidth]{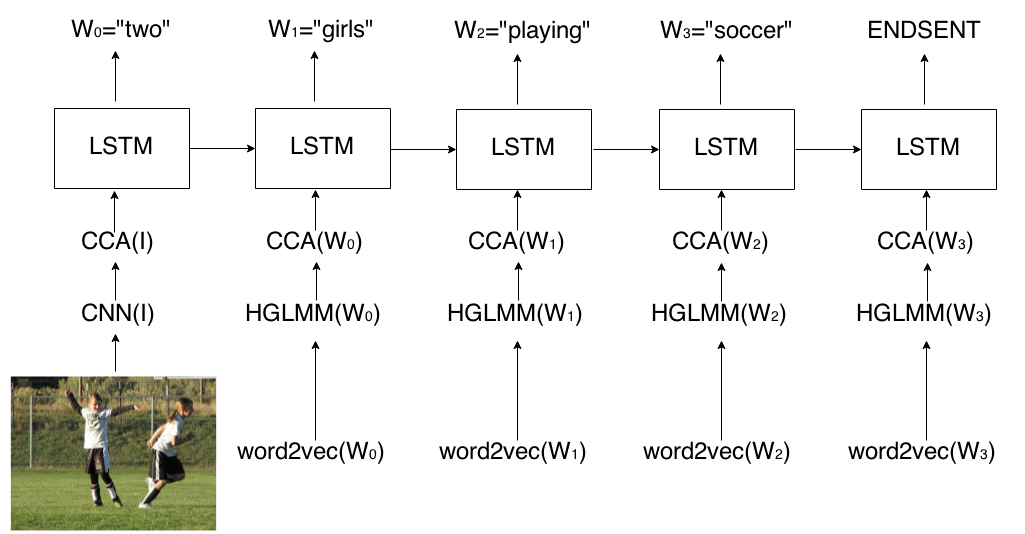}
\caption{\small A running example of our RNN on a sample from Flickr8K. 
The input to the network in the first step is the query image after applying the CNN transformation and then the appropriate CCA projection for images.
The input in every following step $t$ is the word2vec representation of the word that was predicted in step $t-1$ after applying the $HGLMM$ fisher vector representation
and then the appropriate CCA projection for sentences.}
\label{fig:RNNModel}
\end{figure}

In the task of sentence synthesis one is required to generate a novel sentence for a given image query. Specifically, our sentence synthesis model is using a recurrent neural network (RNN) with long short term memory (LSTM) units~\cite{hochreiter1997long}.
Our RNN architecture, shown in Figure~\ref{fig:RNNModel}, is similar to that of~\cite{vinyals2014show}, and is composed of a single LSTM layer with $512$ LSTM units, and a softmax output layer of the size of the dictionary. However, the input of our RNN differs greatly from that of other contributions~\cite{vinyals2014show,karpathy2014deep,mao2014explain,donahue} that are using deep learning for sentence synthesis: Our input is based on the methods and representations that were described previously for the image annotation task. Since both images and words (sentences of length 1) are projected to the same CCA space, our RNN treats the two entities in exactly the same way. Therefore, there is no need to learn an extra layer that projects the words to the image space during the RNN training.

In order to synthesize a sentence, we use a greedy and deterministic algorithm, and do not perform sampling or a beam search as was done in~\cite{vinyals2014show}. This, mainly for computational reasons. At step $t=0$ the image query, $I$, is passed through the VGG~\cite{oxford} deep convolution neural network, producing a vector, $CNN(I)$. 
Applying the CCA image transformation matrix on $CNN(I)$ results with a new vector, $CCA(I)$. The RNN takes $CCA(I)$ as an input and the word, $W_{0}$, that corresponds to the entry with the highest value in the softmax layer is chosen as the first word in the predicted sentence. 

At each step $t>0$, the word that was predicted in the previous step, $W_{t-1}$, is represented according to its word2vec representation, $word2vec(W_{t-1})$. 
Applying the $HGLMM$ fisher vector on $word2vec(W_{t-1})$ results with a new vector, $HGLMM(W_{t-1})$. Finally, the CCA sentence transformation matrix is applied on $HGLMM(W_{t-1})$, producing $CCA(W_{t-1})$. 
The RNN takes $CCA(W_{t-1})$ as an input  and the word, $W_{t}$, that corresponds to the entry with the highest value in the softmax
layer is chosen as the $t$ word in the predicted sentence. This process is repeated until the predicted word becomes {\em ENDSENT} which is a special word that was added to the end of each sentence. 

\begin{figure}
\centering
\includegraphics[width=0.9\linewidth]{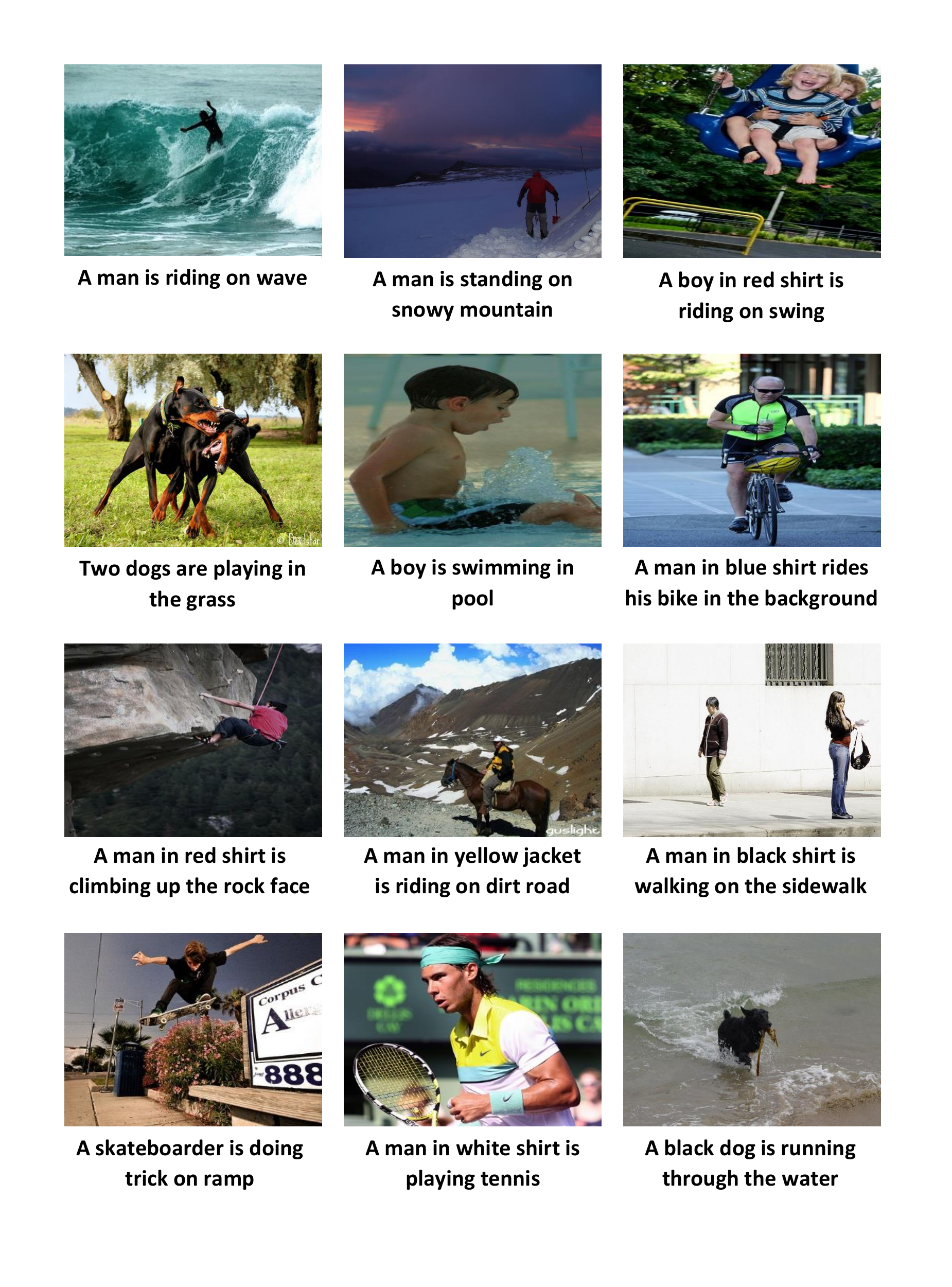}
\caption{\small A few samples from the test set of Flickr 8k and the corresponding sentences that were generated by our RNN sentence synthesis model.}
\label{fig:RNNFlickr8KExamples}
\end{figure}

The RNN model was trained on the Flickr 8k dataset. The train set contains $6091$ images, which are each described by $5$ sentences (or less) that were created by different annotators. 
As mentioned, the special word {\em ENDSENT} was added to the end of each sentence. The training is done using $SGD$ with learning rate of $0.00001$ and a momentum of $0.5$. The model was trained for $300$ epochs where the stopping point was selected
according to the results on the validation set of Flickr 8k. The CURRENNT library ~\cite{weninger2014introducing} was used for training our model. Images from the test set of Flickr 8k and the corresponding sentences that were generated by our RNN sentence synthesis model are presented in Figure ~\ref{fig:RNNFlickr8KExamples}.

\section{Discussion}

The Fisher Vector has proven to be useful in a variety of object recognition applications. In all those applications, it was derived according the Gaussian Mixture Model. In this paper we show that using a Fisher Vector derived from other distributions, namely, LMM and HGLMM, one can obtain improved accuracy in central computer vision tasks. We believe that these improvements would carry on to other tasks as well.

The normalization improvements that were suggested by Perronnin et al.~\cite{perronnin2010large}  dramatically increased the performance achieved by the Fisher Vector technique and contributed to its success. It remains to be explored whether there are specific normalization techniques that are most suitable for the Fisher Vectors that we derive from the LMM and HGLMM distributions.

The Hybrid Gaussian-Laplacian Mixture Model (HGLMM) that we presented, allowed us to gain benefits from both underlying distributions by having the flexibility that each dimension in each component would be modeled according to the most suitable distribution. Such geometric-mean mixtures could be generalized to any two distributions and fit real world data of any distribution shape. It is also not limited to just two parametric distributions.

\section*{Acknowledgments}
This research is supported by the Intel Collaborative Research Institute for Computational Intelligence (ICRI-CI).

{\small
\bibliographystyle{ieee}
\bibliography{hglmm}
}

\end{document}